\documentclass{article}

\usepackage{arxiv}

\usepackage[utf8]{inputenc} 
\usepackage[T1]{fontenc}    
\usepackage{hyperref}       
\usepackage{url}            
\usepackage{booktabs}       
\usepackage{amsfonts}       
\usepackage{nicefrac}       
\usepackage{microtype}      
\usepackage{lipsum}		

\usepackage{graphicx}
\usepackage{tikz}
\usetikzlibrary{calc}
\usetikzlibrary{decorations.pathreplacing,calligraphy}
\usetikzlibrary {arrows.meta}
\usepackage{subcaption}
\usepackage{amsmath}
\usepackage{amssymb}
\usepackage[compact]{titlesec}
\usepackage{enumitem}

\title{PeersimGym: An Environment for Solving the Task Offloading Problem with Reinforcement Learning}

\date{March 2024} 					

\author{ \href{https://orcid.org/0009-0000-9797-9603}{\includegraphics[scale=0.06]{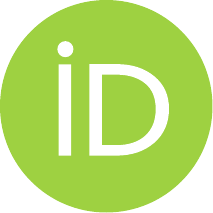}\hspace{1mm}Frederico Metelo} \\
	Department of Computer Sceince\\
	NOVA School of Science and Technology\\
	Caparica, Portugal\\
	\texttt{fc.metelo@campus.fct.unl.pt} \\
    \And
    \href{https://orcid.org/0000-0002-5656-9189}{\includegraphics[scale=0.06]{orcid.pdf}\hspace{1mm}Stevo Racković} \\
	Department of Mathematics\\
	Instituto Superior Técnico\\
	Lisbon, Portugal \\
     \And
	\href{https://orcid.org/0000-0003-3773-3593}{\includegraphics[scale=0.06]{orcid.pdf}\hspace{1mm}Pedro Ákos} \\
	Department of Informatics\\
	NOVA School of Science and Technology\\
	Caparica, Portugal\\	
	\And
	\href{https://orcid.org/0000-0003-3071-6627}{\includegraphics[scale=0.06]{orcid.pdf}\hspace{1mm}Cláudia Soares} \\
	Department of Computer Sceince\\
	NOVA School of Science and Technology\\
	Caparica, Portugal \\	
}

\renewcommand{\shorttitle}{PeersymGym}

\hypersetup{
pdftitle={PeersimGym},
pdfauthor={Frederico Metelo}
}

\begin{document}
\maketitle

\begin{abstract}
Task offloading, crucial for balancing computational loads across devices in networks such as the Internet of Things, poses significant optimization challenges, including minimizing latency and energy usage under strict communication and storage constraints. While traditional optimization falls short in scalability; and heuristic approaches lack in achieving optimal outcomes, Reinforcement Learning (RL) offers a promising avenue by enabling the learning of optimal offloading strategies through iterative interactions. However, the efficacy of RL hinges on access to rich datasets and custom-tailored, realistic training environments.
To address this, we introduce PeersimGym, an open-source, customizable simulation environment tailored for developing and optimizing task offloading strategies within computational networks. PeersimGym supports a wide range of network topologies and computational constraints and integrates a \textit{PettingZoo}-based interface for RL agent deployment in both solo and multi-agent setups. Furthermore, we demonstrate the utility of the environment through experiments with Deep Reinforcement Learning agents, showcasing the potential of RL-based approaches to significantly enhance offloading strategies in distributed computing settings. PeersimGym thus bridges the gap between theoretical RL models and their practical applications, paving the way for advancements in efficient task offloading methodologies.
\end{abstract}


\section{Introduction}

The proliferation of large networks of devices, such as the Internet of Things (IoT), has led to an exponential increase in data generation, requiring significant computational resources for processing. Traditionally, Cloud Computing provided the backbone for such computational demands. However, its limitations in latency and network traffic have become apparent with the growth of device networks~\cite{min2019learning}. Edge Computing emerged as a paradigm shift, extending the Cloud to bring processing capabilities closer to data sources, mitigating latency and traffic issues.
In this evolving landscape, task offloading---the distribution of computational tasks across network participants---has gained prominence, particularly within the Fog and Multi-access Edge Computing (MEC) paradigms. Both paradigms aim to decentralize computing power, bringing it closer to end-users and alleviating the constraints of traditional Cloud Computing~\cite{Muniswamaiah_Manoj_2021,Varghese_Buyya_2018}. 
Despite distinctions between MEC and Fog computing, this paper treats them interchangeably, focusing on their shared goal of minimizing device-to-cloud distances~\cite{yu2020deep}. Addressing the challenges of task offloading in such distributed environments involves balancing numerous factors, including task latency, energy consumption, and task completion reliability~\cite{zhu2019blot}.
Conventional optimization methods often struggle to efficiently manage these complex systems. Multi-Agent Reinforcement Learning (MARL) offers promise in optimizing resource allocation and scheduling to maximize system efficiency or meet specific performance metrics~\cite{lin2023deep,zhang2023cooperative}. In the task offloading problem, a reinforcement learning (RL) agent can serve as a decision maker to learn the optimal task-resource allocation strategy by interacting with the environment (i.e., the task and available resources). However, it is impractical to train agents in the real world. Therefore, a standardized, customizable, efficient, and user-friendly simulation tool is essential. 

\textbf{Contributions.}
Our contributions focus on advancing the field of MARL and Edge computing through the introduction of the PeersimGym environment and comprehensive experimental analysis. The \textbf{PeersimGym Environment} is a highly adaptable simulation platform tailored for the development, training, and evaluation of MARL strategies for task offloading challenges in Edge Computing systems. This environment, built on the synergy between an edge system simulator built with Peersim P2P simulator~\cite{peersim} and the PettingZoo API~\cite{PettingZoo_2020}, allows detailed configuration of network topologies, node characteristics, and task parameters, facilitating a wide range of experimental setups. An \textbf{experimental analysis} demonstrates the capability of the PeersimGym to train effective MARL solutions, exemplified by the performance of a Double Deep Q Network and Advantage Actor Critic algorithm. Our analysis compares the performance of DRL agents to several non-RL algorithms across various network configurations and task offloading scenarios, highlighting the advantages of RL approaches in optimizing Edge Computing networks.
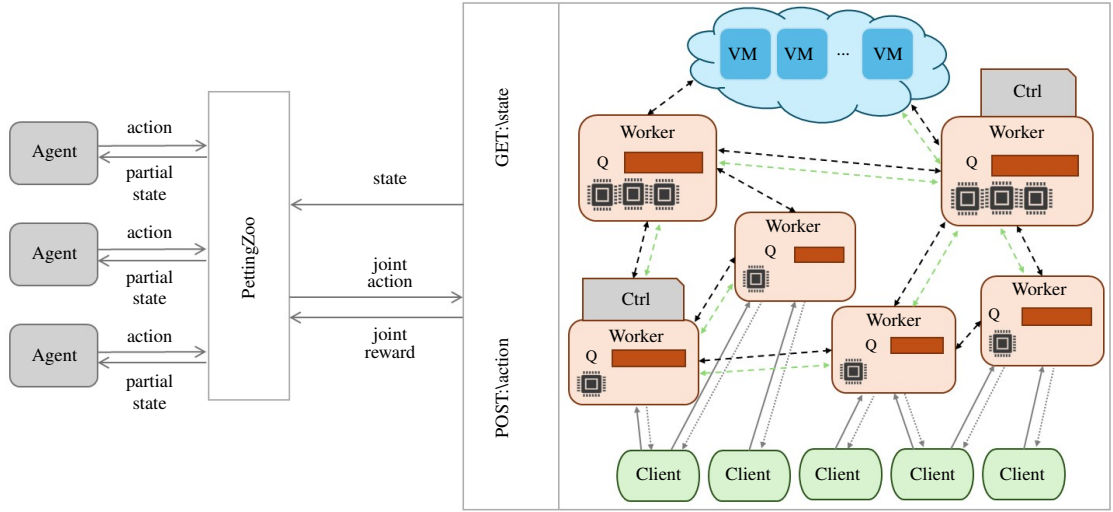
\begin{figure}[t]
\centering
\begin{tikzpicture}
    \node[above right, inner sep=0] (image) at (0,0)
    {\includegraphics[width=0.9\linewidth]{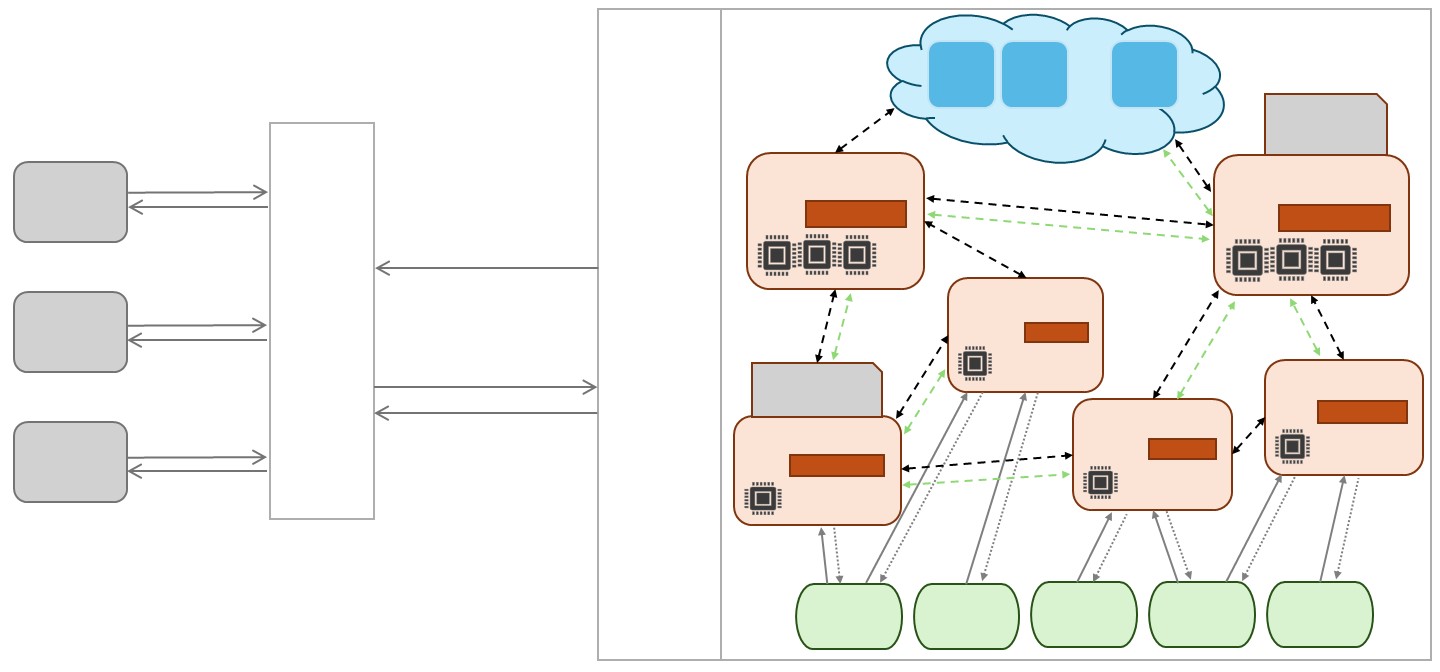}};
    \begin{scope}[
        x={($0.1*(image.south east)$)},
        y={($0.1*(image.north west)$)}]
        \node[black] at (5.90,0.85) {\scriptsize Client};
        \node[black] at (6.70,0.85) {\scriptsize Client};
        \node[black] at (7.50,0.85) {\scriptsize Client};
        \node[black] at (8.35,0.85) {\scriptsize Client};
        \node[black] at (9.15,0.85) {\scriptsize Client};
        \node[black] at (5.7,3.5) {\scriptsize Worker};
        \node[black] at (5.8,7.4) {\scriptsize Worker};
        \node[black] at (7.1,5.6) {\scriptsize Worker};
        \node[black] at (8.0,3.8) {\scriptsize Worker};
        \node[black] at (9.3,4.4) {\scriptsize Worker};
        \node[black] at (9.1,7.4) {\scriptsize Worker};
        \node[black] at (9.2,8.2) {\scriptsize Ctrl};
        \node[black] at (5.7,4.2) {\scriptsize Ctrl};
        \node[black] at (5.3,3.1) {\tiny Q};
        \node[black] at (5.4,6.8) {\tiny Q};
        \node[black] at (6.9,5.1) {\tiny Q};
        \node[black] at (7.8,3.3) {\tiny Q};
        \node[black] at (8.9,3.9) {\tiny Q};
        \node[black] at (8.7,6.8) {\tiny Q};
        \node[black] at (6.65,8.9) {\scriptsize VM};
        \node[black] at (7.15,8.9) {\scriptsize VM};
        \node[black] at (7.95,8.9) {\scriptsize VM};
        \node[black] at (7.55,8.9) {\scriptsize ...};
        \node[black, rotate=90] at (4.5,2.5)  {\scriptsize POST:\textbackslash action};
        \node[black, rotate=90] at (4.5,7.5)  {\scriptsize GET:\textbackslash state};
        \node[black, rotate=90] at (2.25,5)  {\scriptsize PettingZoo};
        \node[black] at (0.5,7) {\scriptsize Agent};
        \node[black] at (0.5,5.05) {\scriptsize Agent};
        \node[black] at (0.5,3.1) {\scriptsize Agent};
        \node[black] at (3.5,6.5) {\scriptsize state};
        \node[black] at (3.5,4.85) {\scriptsize joint};
        \node[black] at (3.5,4.55) {\scriptsize action};
        \node[black] at (3.5,3.5) {\scriptsize joint};
        \node[black] at (3.5,3.2) {\scriptsize reward};
        \node[black] at (1.35,7.5) {\scriptsize action};
        \node[black] at (1.35,6.6) {\scriptsize partial};
        \node[black] at (1.35,6.2) {\scriptsize state};
        \node[black] at (1.35,5.5) {\scriptsize action};
        \node[black] at (1.35,4.6) {\scriptsize partial};
        \node[black] at (1.35,4.2) {\scriptsize state};
        \node[black] at (1.35,3.5) {\scriptsize action};
        \node[black] at (1.35,2.6) {\scriptsize partial};
        \node[black] at (1.35,2.2) {\scriptsize state};
    \end{scope} 
    \end{tikzpicture}
    \caption{Simulation Pipeline Overview. Left: PettingZoo API integration, facilitating agent-simulation interaction via Python and RESTful requests for practical task offloading optimization. Right: network topology with worker nodes and connections, enabling task generation and state sharing for RL agent training. }
    \label{fig:FullDiagram}
\end{figure}
The \textbf{source code} for PeersimGym, along with usage documentation and testing resources, is made available in the \href{https://github.com/FredericoMetelo/peersim-environment}{Simulator repository}\footnote{\href{https://github.com/FredericoMetelo/peersim-environment}{https://github.com/FredericoMetelo/peersim-environment}}, and the \href{https://github.com/FredericoMetelo/TaskOffloadingAgentLibrary}{Agent repository}\footnote{\href{https://github.com/FredericoMetelo/TaskOffloadingAgentLibrary}{https://github.com/FredericoMetelo/TaskOffloadingAgentLibrary}}, fostering further research and community engagement in advancing Edge Computing solutions.

\section{Background and Related Work}

Task offloading addresses the redistribution of computationally intensive tasks from resource-limited devices to more capable ones to enhance system performance. This process requires strategic decision-making regarding the \textit{what}, \textit{where}, \textit{how}, and \textit{when} aspects of offloading tasks. Literature on task offloading distinguishes vertical offloading to higher-tier systems~\cite{qiu2019online}, horizontal offloading among peers~\cite{zhu2019blot,baek2019managing}, and hybrid approaches that blend both directions~\cite{Baek22Fog}.
The choice of offloading destinations varies significantly, considering idle nodes, those with shorter task queues~\cite{baek2019managing}, proximity constraints~\cite{van2018deep,yu2020deep}, or unrestricted selection accounting for potential offloading failure consequences~\cite{Baek22Fog}. Failures in offloading can arise from latency constraints \cite{dai2022task}, node capacity limitations~\cite{van2018deep}, energy shortages, or other~\cite{peng2022deep}.
Task modeling further diversifies the field, encompassing homogeneous versus heterogeneous tasks~\cite{liu2019deep}, divisible versus indivisible tasks \cite{min2019learning}, and dependent versus independent tasks~\cite{geng2023deep}. Common objectives for task offloading include minimizing computation and transmission latency~\cite{yu2020deep}, reducing energy consumption~\cite{huang2021deadline}, and avoiding task failures~\cite{min2019learning,baek2019managing}. Other considerations such as task utility~\cite{jain2023qos}, queue wait times, and CPU utilization~\cite{yu2020deep} are less frequent.
Our simulator, PeersimGym, is designed to accommodate this broad spectrum of offloading strategies, task models, and objectives. It offers the flexibility to configure various aspects of the simulation environment, enabling the exploration of a wide range of scenarios and contributing to a more nuanced understanding of task offloading dynamics in distributed computing networks.

\textbf{Reinforcement Learning (RL) for Task Offloading.}
RL is a powerful and dominant approach for solving the task offloading problem, as it can find an optimal solution with excellent efficiency, given a well-defined environment and correct reward shaping. RL has been applied to various Fog and MEC settings, considering single agent methods~\cite{van2018deep}, as well as multi-agent with a set of independent learners~\cite{dai2022task} or in federated cooperation~\cite{tong2023multi}. Each category assumes different observability and sharing among nodes. Models can be fully or partially observable~\cite{Baek22Fog}, with local or global optimization objectives in a multi-agent case. Learning agents range from tabular methods~\cite{baek2019managing}, and multi-armed bandits~\cite{zhu2019blot}, to complex deep Q networks~\cite{tong2023multi} or actor-critic agents~\cite{Baek22Fog}. 

Our contribution addresses a gap in the existing literature by introducing a training environment for agents, facilitating uniform comparisons across different solutions. Given the diverse nature of these solutions, our simulation tool offers a high level of customizability to accommodate the diverse requirements. Specifically, we provide a simulation platform tailored for training both centralized and decentralized reinforcement learning algorithms, targeting task offloading in edge systems.
The proposed simulator includes a PettingZoo~\cite{PettingZoo_2020} environment to interface the simulation as an integral component that allows a user to design and train an arbitrary RL agent(s) over a selected environment.

\textbf{Comparative advantages of PeersimGym over existing simulators and environments.}
Available simulators for edge-like networks~\cite{Somnez_2018_EdgeCloudSim,Redowan_Mahmud_2021_ifogsim2} are not prepared out of the box for RL training and do not provide a high level of flexibility for different protocols and topologies. There are also environments for RL that allow for training task offloading RL agents~\cite{Boni_Hassan_Drira_2021_ns3gym,Santos_Wauters_2020_FogGym}; however, the solution of~\cite{Boni_Hassan_Drira_2021_ns3gym} requires implementing the routing mechanism for multi-agent reinforcement learning, and of~\cite{Santos_Wauters_2020_FogGym} is built on the engine from~\cite{Boni_Hassan_Drira_2021_ns3gym}, and focuses only on a single task vertical offloading scenario. PeersimGym addresses these limitations by enabling the user to configure multiple task offloading scenarios, namely that of horizontal task offloading in the P2P setting for Load Balancing. Furthermore, PeersimGym uses PettingZoo, a version of OpenAI gym~\cite{PettingZoo_2020} focused on MARL, and it provides an API better adapted for the task~\cite{PettingZoo_2020} than OpenAI gym and its successor Gymnasium~\cite{towers_gymnasium_2023}. To the best of our knowledge, PeersimGym is the only environment for RL developed with MARL task offloading and high configurability and modality as its central focus.

\section{Enhancing Task Offloading with MARL}\label{sec:RLenvironmentComponent}

To address the complexity of Task Offloading in highly complex edge environments, researchers leverage RL and its subset, DRL, and by exploring the distributed nature of these systems MARL emerges as promising solution.
As a tool to develop MARL algorithms, PeersimGym incorporates a Python-based environment tailored for developing, training, and deploying RL models for task offloading, aligned with the \textit{PettingZoo} framework.
With the set of experiments in this paper, we showcase how our simulator can enable further RL and DRL contributions to this field in Sec.~\ref{sec:Exp}.
Certain Nodes within the network act as RL agents. These agents have the capability to observe the state (albeit partially) and make informed decisions regarding whether to process tasks locally or offload them. This decision-making process is influenced by the need to balance between local processing, the risk of queue overflow, and the costs associated with task offloading, including potential overload of other nodes.

\begin{figure}[t]
\begin{subfigure}[c]{0.4\textwidth}
{
\begin{tikzpicture}
    \node[above right, inner sep=0] (image) at (0,0)
    {\includegraphics[width=\textwidth]{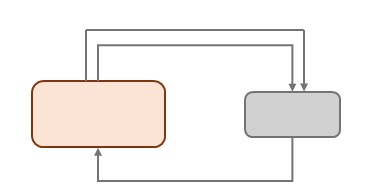}};
    \begin{scope}[
        x={($0.1*(image.south east)$)},
        y={($0.1*(image.north west)$)}]
            \node[black] at (2.6,4) {\scriptsize Environment};
            \node[black] at (7.75,4) {\scriptsize Agent};
            \node[black] at (5.3,1.25) { \scriptsize action $a_t$};
            \node[black] at (5.3,6.9) {\scriptsize reward $r_t$};
            \node[black] at (5.3,9) {\scriptsize state $s_t$};
        \end{scope}
    \end{tikzpicture}
    }
    \caption{Reinforcement Learning Loop.}
    \end{subfigure}
\begin{subfigure}[c]{0.575\textwidth}
{
\begin{tikzpicture}
    \node[above right, inner sep=0] (image) at (0,0)
    {\includegraphics[width=\textwidth,trim={0 0.9cm 0 0},clip]{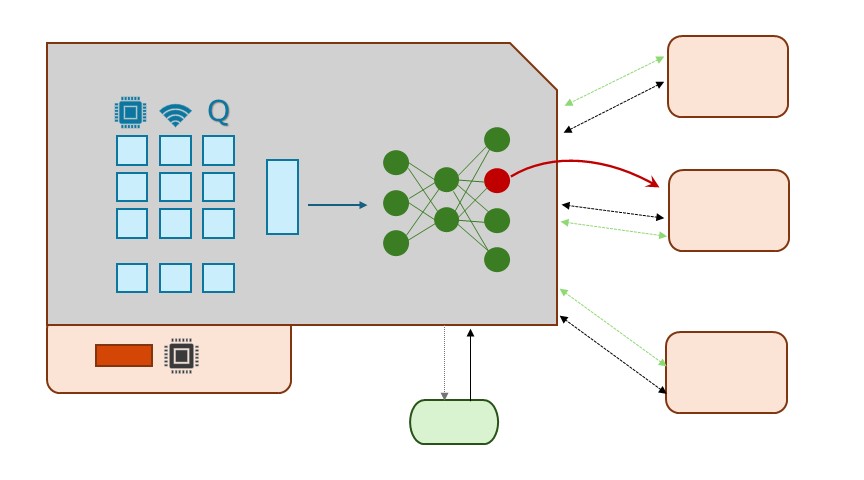}};
    \begin{scope}[
        x={($0.1*(image.south east)$)},
        y={($0.1*(image.north west)$)}]
        \node[black] at (2,8.5) {\scriptsize Observation $s_t$};
        \node[black] at (2,4.4) {\scriptsize +};
        \node[black] at (2.9,5.8) {\scriptsize +};
        \node[black] at (3.3,6.8) {\tiny tasks};
        \node[black] at (1.1,3.6) {\tiny w\textsubscript{0}};
        \node[black] at (1.1,5.) {\tiny w\textsubscript{1}};
        \node[black] at (1.1,5.7) {\tiny w\textsubscript{2}};
        \node[black] at (1.1,6.5) {\tiny w\textsubscript{3}};
        \node[black] at (5,7.2) {\scriptsize DQN};
        \node[black] at (7.40,6.8) {\scriptsize Action $a_t$};
        \node[black] at (3.00,2.0) {\scriptsize w\textsubscript{0}};
        \node[black] at (8.50,1.8) {\scriptsize w\textsubscript{1}};
        \node[black] at (8.50,5.10) {\scriptsize w\textsubscript{2}};
        \node[black] at (8.50,8.25) {\scriptsize w\textsubscript{3}};
        \node[black] at (3,9.5) {\scriptsize Controller Protocol};
    \end{scope}
    \end{tikzpicture}
    }
    \caption{Local state and action for a worker node.}
    \end{subfigure}
    \caption{General and problem-specific RL state action overview.}
    \label{fig:StateAction}
\end{figure}
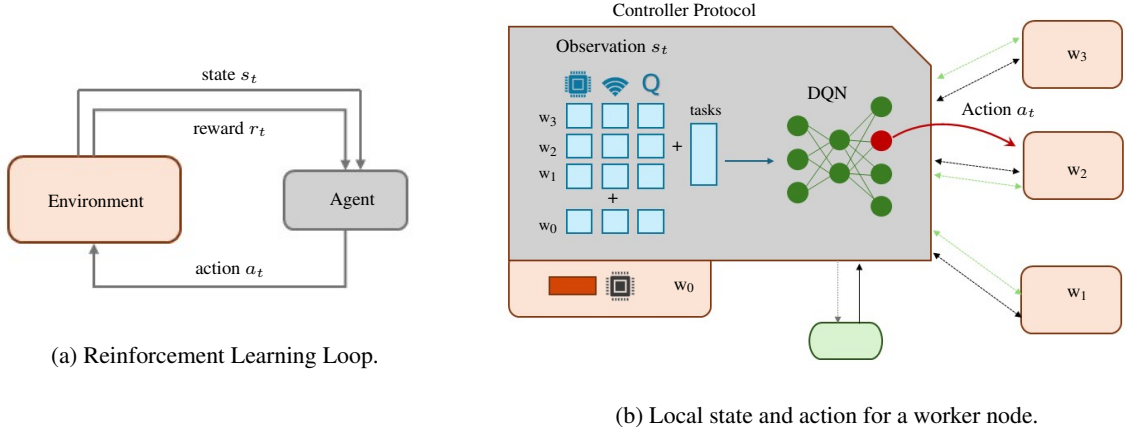

\textbf{The role of Reinforcement Learning Agents.} 
The interaction cycle of an RL agent with its environment is structured around a continuous loop where, at each timestep $t=1,...,T$, the agent observes the system state $s_t$, executes an action $a_t$ based on this observation, and receives feedback in the form of a reward $r_t$.
This feedback reflects the effectiveness of the action, taking into account both its immediate impact and its influence on future states. Through this iterative process, the agent refines its policy---a set of rules determining its actions in various states---to maximize cumulative rewards, thereby aligning with the goal of optimizing task offloading decisions (Fig.~\ref{fig:StateAction}).

\textbf{Deep Reinforcement Learning for Task Offloading.}
Deep Q Networks (DQN)~\cite{van2018deep} and their variants, such as Double Deep Q Networks (DDQN)~\cite{mnih2015human} and Actor-Critic methods, like the Advantage Actor-Critic (A2C)\cite{Baek22Fog}, are prominent DRL approaches applied to task offloading. These methodologies have been shown to stabilize training and enhance learning efficacy through sophisticated neural network architectures that approximate optimal action-selection policies.
In PeersimGym, the flexibility of our RL environment supports the integration of various DRL models. For illustrative purposes, we focus on implementations of DDQN and A2C, reflecting their proven effectiveness in recent literature~\cite{dai2022task,Baek22Fog}. This choice underscores the potential of DRL to address the complexities associated with task offloading in Edge Computing, as shown in Sec.~\ref{sec:Exp}. 

\textbf{Framework and Model Dynamics.}
Task offloading decisions within our simulated environment are modeled as a Markov Game (MG), accommodating the multi-agent aspect of Edge Computing networks. This formulation extends the Markov Decision Process (MDP) framework to scenarios involving multiple decision-makers, thereby capturing the interactive and competitive nature of task offloading.
Single-agent RL is modeled as an MDP, with a sequence of states such that the Markov Property holds, i.e., the next state $s_{t+1}$ depends exclusively on the current state $s_t$ and the performed action $a_t$. When including multiple agents, most MDP convergence properties do not hold, hence, we formulate our problem as an MG~\cite{Nowé2012}. In our setting, in a network of $N$ nodes, an MG is represented as a tuple $\langle n, \mathcal{S}, \mathcal{A}, P, R, \gamma \rangle$, where $n$ is the number of agents (nodes with a controller protocol), $\mathcal{S}$ is the state space, $\mathcal{A}$ is the action space, $P$ is an (unknown) transition probability function, $R$ is the reward function, and $\gamma$ is a reward discount factor. Next, we introduce these constructs in more detail.

\textbf{State Space.} 
The state space in our environment is designed to be highly customizable, enabling the modeling of various Edge Computing scenarios. 
At time step $t$, each node will broadcast its local state to each of its neighbors and receive their local states to build a local state representation before deciding on the action. The state space is represented by a tuple $\mathcal{S}=(\mathcal{I}, \mathcal{K}, \mathcal{Q}^t,\mathcal{F}, \mathcal{L},\{\mathcal{B}_1,...,\mathcal{B}_n\},\\ \{\mathcal{P}_1,...,\mathcal{P}_n\})$, with elements 
\begin{itemize}
    \item $\mathcal{I}=\{1,...,n\}$, array with the IDs of all the nodes in the network;
    \item $\mathcal{K}=\{\kappa_1,...,\kappa_n\}$, layer/tier for each node;
    \item $\mathcal{Q}^t=\{Q^t_1,..., Q^t_n\}$, queue size for each node in the network, at time step $t$;
    \item $\mathcal Q_i^{\text{max}}=\{Q_1^{\text{max}},..., Q_n^{\text{max}}\}$, maximum capacity for each node's queue;
    \item $\mathcal{N}_n$, IDs of the nodes in node $n$'s neighborhood;
    \item $\mathcal{F}=\{N_{\phi}^1\phi_1,...,N_{\phi}^n\phi_n\}$, processing power for each node;
    \item $\mathcal{L}=\{l_1,...,l_n\}$, position for each node in the network;
    \item $\mathcal{B}_i=\{B_{i,1},...,B_{i,n}\}$, channels' bandwidth for node $w_i$ to all its neighbors;
    \item $P_i$, the transmission power of  node $w_i$’s antenna.
\end{itemize}
The size of the state vector will define the input size of a DRL agent (Fig.~\ref{fig:StateAction} (b)); hence, in this version of the simulator, the state dimension needs to stay consistent throughout iterations.

\textbf{Action Space.}
The action space is similarly designed to reflect the decision-making process regarding task offloading, with actions representing the choice of offloading destinations. This setup facilitates the exploration of strategies that balance local processing advantages against the costs and implications of offloading.
The action space, $\mathcal{A}$, corresponds to the output layer of a DRL agent (see Fig.~\ref{fig:StateAction}). The action $a_t\in\{1,...,N\}$ represents the index of the destination node, which might be one of the neighboring workers or the observed node itself, in case it decides to process the task locally. We use $a_t$ interchangeably to denote both the index of the destination node and the act of sending a task to that node, as long as clarity is maintained within the context. 

\textbf{Reward Function.}
The reward function is a critical component guiding the learning process of the RL agent. It is constructed to reward actions that enhance utility---such as task completions---while penalizing undesirable outcomes like excessive delays or system overloads. This balance encourages the development of nuanced offloading policies that consider various operational constraints and performance metrics. By incorporating reward shaping~\cite{ng1999rewardshaping}, we further refine the learning process, enabling agents to navigate the complex decision space of task offloading more effectively. This approach not only facilitates faster convergence to optimal policies but also allows for a more nuanced understanding of the trade-offs inherent in Edge Computing task management.
The reward shaping term can be defined as $F(s_t, a_t, s_{t+1}) = \Xi(s_{t+1}) - \Xi(s_t)$, where $\Xi(s_t)$ is a user-defined potential function over a state, $s_t$ such that the reward becomes $R' = R + F$.

In this paper, we adopt a reward function as defined in~\cite{baek2019managing}, i.e., a reward function for agent $w_i$, $R_i$, is structured to maximize the utility, $U_i(s_t, a_t)$, and minimize the total delay, $D_i(s_t, a_t)$, and the overloading cost, $O_i(s_t, a_t)$. In particular, the reward for the action $a_t$ in state $s_t$, received by an agent $w_i$ by offloading (or not) task $\tau_k$, is given by
\begin{equation} 
\label{eq:rwrdFunciton}
    R_i(s_t, a_t) = r_u - (D_i(s_t, a_t) + \chi_OO_i(s_t, a_t)),
\end{equation} where $r_u$ is a utility reward and represents the gain over completed tasks, and $\chi_O$ is overloading cost weight. Each term of the reward function is explained in detail below. First we introduce an indicator function, $I_i(a_t)$, such that $I_i(a_t)=1$ \textit{iff} $a_t=i$, meaning the task is meant to be processed in node $w_i$, otherwise $I_i(a_t)=0$.
The reward function for an agent in the context of task offloading is composed of two primary components: the delay function and the cost of overloading. Each incorporates specific parameters and equations that encapsulate the complexities of decision-making in edge-computing environments.
The delay function denoted as $D_i(s_t, a_t)$, is a comprehensive measure that accounts for three distinct temporal aspects associated with task offloading, namely
\begin{equation}
D_i(s_t, a_t) = \chi_D^{\text{wait}} T_{i,a_t}^{\text{wait}}(\tau_k) + \chi_D^{\text{comm}} T_{i,a_t}^{\text{comm}}(\alpha_k^{\text{out}}) + \chi_D^{\text{exc}} T_{i,a_t}^{\text{exc}}(\tau_k),
\end{equation}
Here, $\chi_D^{\text{wait}}$, $\chi_D^{\text{comm}}$, and $\chi_D^{\text{exc}}$ serve as hyperparameters, adjusting the weight of each time-related component within the overall delay function. The waiting time $T_{i,a_t}^{\text{wait}}(\tau_k)$ reflects the duration a task $\tau_k$ spends in the queue, either at the originating node $w_i$ or at an offloading target $w_j$, and is given by 
\begin{equation}
T_{i,a_t}^{\text{wait}}(\tau_k) = \frac{Q_i^t}{N^i_{\phi}\phi_i} + \sum_{j\neq i}\frac{Q_j}{N^j_{\phi}\phi_j} I_j(a_t),
\end{equation}
where $Q_i^t$ is the queue size at time $t$, $N^i_{\phi}$ and $\phi_i$ represent the number of processors and their frequency at node $w_i$, respectively.
To quantify the transmission efficiency, specifically the rate at which bits are communicated per second, we invoke the \textit{Shannon-Hartley} theorem~\cite{book:Anttalainen_2003}. Accordingly, the communication latency for transmitting $\alpha_k^{\text{out}}$ bits between nodes $w_i$ and $w_{a_t}$ is given by
\begin{equation}\label{eq:time_communication}
T^{\text{comm}}_{i,a_t}(\alpha_k^{\text{out}}) = \frac{\alpha_k^{\text{out}}}{B_{i,a_t} \log(1 + 10^{\frac{P_i + G_{i,a_t} - \omega_0}{10}})},
\end{equation}
where $B_{i,a_t}$ is the bandwidth of the communication channel between nodes $w_i$ and $w_{a_t}$, $P_i$ denotes the transmission power of the source node $w_i$, $G_{i,a_t}$ is the channel gain, and $\omega_0$ represents the noise power in the communication channel. This formulation underpins our model for evaluating the communication overhead associated with task offloading in Edge networks. $P_i$, $G_{i,a_t}$ and $\omega_0$ are measured in dB.
The execution cost difference $T_{i,a_t}^{\text{exc}}(\tau_k)$ between local and target node processing is quantified as
\begin{equation}
T_{i,a_t}^{\text{exc}}(\tau_k)= \frac{\rho_k \xi_k}{N_{\phi}^{a_t}\phi_{a_t}} - \frac{\rho_k \xi_k}{N_{\phi}^{i}\phi_{i}},
\end{equation}
indicating the variation in processing time due to differences in node capabilities.
The cost of overloading, $O_i(s_t, a_t)$, emphasizes the potential system strain caused by task offloading, expressed as $O_i(s_t, a_t) = -\log (p_t^{{a_t}})/3$. The probability of overloading $p_t^{{a_t}}$ and the expected queue state $Q'_{a_t}$ are critical in assessing the impact of offloading decisions on the target node's workload, calculated as $p_t^{{a_t}} = \max \left(0, \frac{Q_{a_t}^{\max} - Q_{a_t}}{Q_{a_t}^{\max}}\right),$ and $Q'_{a_t} = \min( \max(0, Q_{a_t} - \phi_{a_t}) + 1, Q_{a_t}^{\text{max}})$, where $Q_{a_t}^{\text{max}}$ is the maximum queue capacity of the offloading target, and $\phi_{a_t}$ is its processing rate. These equations collectively frame the decision-making landscape for RL agents, highlighting the intricate balance between task processing efficiency, communication overheads, and system resilience against overloading.

\section{PeersimGym}\label{sec:Method} 
PeersimGym introduces a novel framework for simulating and training task-offloading MARL algorithms in Edge networks. We leverage the versatile Java-based PeerSim P2P simulator~\cite{peersim} and extend it to model edge systems. Additionally, PeersimGym incorporates a Python API compatible with the PettingZoo framework~\cite{towers_gymnasium_2023}, offering an intuitive structure for developing RL for task offloading. This section delineates the two primary components of our tool: a simulator for custom system network creation and a Python environment for constructing and training RL models, grounded in PettingZoo principles (Fig.~\ref{fig:FullDiagram}).

\subsection{System Modeling}\label{sec:SystemModel}
PeersimGym allows a high level of customization in crafting edge systems, enabling customization of components and their attributes.
We introduce models for nodes, tasks, and communication, which collectively define the communication dynamics of the simulation.

\textbf{Node Model.}
The simulation framework models a network comprising client devices $\mathcal{C}$, akin to IoT sensors, which generate and dispatch data for processing. This data is handled by worker nodes, which can either process tasks locally or offload them to other nodes with available resources. Worker nodes $\mathcal{W}$ possess distinct characteristics, including:
\begin{itemize}
    \item \textbf{Task queue}, $Q_i$, a data structure that allows at most $Q_i^{\text{max}}$ received tasks to be stored and await to be processed in a first-in-first-out fashion. Any tasks received above the capacity of the node will be dropped.
    \item \textbf{Number of processors}, $N_{\phi}^i$, of frequency, $\phi_i$. The node can process $N_{\phi}^i\phi_i$ instructions per time step.
    \item \textbf{Transmission power}, $P_i$, that affects the wireless communication delays.
    \item \textbf{Location of the node}, $l_i$, which also affects the communication delay and other proximity-based mechanisms.
\end{itemize} 
Worker nodes $\mathcal{W}$ are categorized into tiers, reflecting a hierarchy similar to the fog computing model. This tiered structure, alongside the optional integration of Cloud servers, facilitates the modeling of various network architectures from P2P to hierarchical n-tier systems (Fig.~\ref{fig:FullDiagram}). Nodes containing an RL agent can offload tasks to their neighbors, managing offloading decisions and maintaining state information of adjacent nodes.

\textbf{Task Model.}
 The workload originates from processing tasks $\tau_i$, denoted as tuples representing computational demands. These tasks, generated by clients, include attributes such as instruction count, input/output data sizes, CPU cycle cost per instruction, and processing deadlines. Specifically, a task is represented as
$\tau_i = \langle i, \rho_i, \alpha_i^{\text{in}}, \alpha_i^{\text{out}}, \xi_i, \delta_i \rangle$, where
     $i$ is a unique identifier of the task;
     $\rho_i$ is the number of instructions to be processed;
     $\alpha_i^{\text{in}}$ is the total data size of the input;
     $\alpha_i^{\text{out}}$ is the data size of the output/results;
     $\xi_i$ is the cost in CPU cycles per instruction;
     $\delta_i$ is the deadline of the task, i.e., the maximum allowed latency for returning the results.
 Task arrival follows a Poisson distribution, and if capacity is exceeded, tasks are dropped (Fig.~\ref{fig:FullDiagram}).
 
\textbf{Communication Model.}
Task offloading and reception are simulated under the assumption of a generic wireless communication model, as this is the most common in the literature.  Utilizing the Shannon-Hartley theorem~\cite{book:Anttalainen_2003}, we calculate the bits transmitted per second, incorporating factors such as channel bandwidth, channel gain, and transmission power (see Eq. \ref{eq:time_communication}). 

In the context of task generation and reception, nodes equipped with a \textit{controller protocol} play a pivotal role in determining whether to process tasks locally or offload them. This decision-making process is facilitated through the exchange of local state information via one-hop broadcasts. Communication within the network, particularly the offloading interactions among nodes, is predicated on wireless transmission. 

\subsection{Implementation details}

\textbf{PeersimEnv.}
PeersimGym integrates the Peersim simulator and the PettingZoo environment (PZ env), facilitating the configuration and execution of simulations through REST requests. This integration is enabled by encapsulating the Peersim simulation within a Spring Boot REST server, providing endpoints for action posting and state retrieval, adhering to PettingZoo standards (Fig.~\ref{fig:FullDiagram}).  A full list of the available configurations can be found in the code repository.

\textbf{Simulator.}
At its core, PeersimGym employs the Peersim simulation tool, supporting both event-driven and cycle-driven engines. 
The simulation models a network as a collection of nodes running various protocols, including client, worker, controller, and a customized Simulation Manager protocol. This Simulation Manager periodically pauses the simulation to process actions and resume, maintaining a cyclical operation that aligns with the specified behaviors of worker, client, and controller protocols.
Through extensive configurability and a focus on customization, PeersimGym presents a powerful tool for simulating Edge network environments and training RL models for task offloading, allowing for more sophisticated and realistic simulations in Edge Computing research.

\textbf{Enhanced Simulation Protocols and Event Handling.}
In the simulation environment of our system, each node adheres to specific protocols designed to mimic the behavior of Edge Computing networks accurately. These protocols, namely the worker, client, and controller protocols, are essential for the dynamic interaction between nodes, ensuring the efficient processing and distribution of tasks. Below, we outline the refined functionalities of these protocols.
\begin{itemize}
\item \textbf{Worker protocol:} 
Central to our simulation, the worker protocol governs the processing mechanics of tasks within the network. Upon receiving a task, the worker updates a task-specific instruction counter to track its progress. Completion of a task triggers the protocol to return the result to the originating client. This return path may involve multiple hops, particularly for tasks that have been offloaded across several nodes.  Should the queue of the worker deplete, it transitions into an idle state, awaiting new tasks.
\item \textbf{Client protocol:} 
The client protocol is responsible for task distribution within the network. It employs a Poisson process to determine the allocation of tasks to neighbors, with eligibility for task receipt specified in the system configuration. This probabilistic approach ensures a realistic simulation of task dissemination behavior observed in Edge Computing scenarios.
\item \textbf{Controller protocol:} 
This protocol is coupled with a worker protocol, and monitors the worker's state. Upon detecting a state alteration, the controller initiates a one-hop broadcast to disseminate the updated state information to neighboring nodes. Maintaining an updated network state and facilitating informed decision-making for task offloading. It also acts as the bridge between the simulation and the RL agents, passing the offload instructions to the worker, virtually representing the agents, we shall refer to the offload decisions as if they are made by the Controller.
\end{itemize}

\textbf{Event Handling Mechanisms.}
To enhance the fidelity of our simulation, we implement a realistic event-handling system. This system allows each protocol to respond to specific events that reflect real-world interactions within an Edge Computing environment. These events include the following:
\begin{itemize}
    \item \textbf{Worker Protocol Events:} Engages in routines for handling offloaded tasks from other nodes, new tasks directly from clients, and results of concluded tasks. If space permits, tasks are added to the queue for processing; otherwise, they are dropped. Completed tasks are either directed back to the origin client or offloaded to closer workers for final delivery.
    \item \textbf{Client Protocol Event:} Manages the completion of tasks, focusing on registering relevant metrics upon the conclusion of tasks sent to workers.
    \item \textbf{Controller Protocol Event:} Handles the Neighbour State update event, ensuring the node updates its information regarding the state of neighboring nodes as necessary.
\end{itemize}

In the case of the controller protocol, the offloading event is different from the other events. The instructions are received through a REST request from the simulator and are passed by a special class evoking a method directly on each of the controllers which will take the actions immediately

Therefore, by implementing Java classes extending the \textit{AbstractWorker}, \textit{AbstractClient}, and \textit{AbstractController} provided, the user can change the behavior of the elements in the simulation. Furthermore, each of the simulations can be configured based on configuration files. We allow an in-depth configuration of most aspects of the simulation, which include the network topology, the manipulation of what nodes support the controller functions, most properties of the tasks, and the configuration of all protocols. We have also developed a configuration helper tool to simplify the creation of the configuration files. The documentation for all the possible configurations can be found in the environment repository. 

The main focus of the introduced simulator is on being highly customizable, which includes implementing new protocols that make use of the highly innate modality of the Peersim Tool and the extensions we put in place to manage the simulation of Edge Networks. To tailor the simulation to the required scenario, it is possible to define different protocols for the clients, controllers, and workers, that extend their abstract implementations. Furthermore, different actions and information available to each node can be defined, and the communication model and neighborhood definition can also be customized. We provide multiple base classes that allow doing out-of-the-box, binary task offloading, where clients generate indivisible tasks and controllers make decisions on where to offload the full tasks; or batch binary task offloading, where clients generate indivisible tasks but controllers decide for each of the tasks arrived in the time after the last offloading decision where they should be processed. More information on what each one does, how to load different modes, and how to create other implementations can be found on the simulator repository. In this paper, we focus on the binary task offloading implementations.

\textbf{Simulation Data.}
PeersimGym provides a set of resources for collecting data from the simulation that help with the development process of agents and to provide insights on the behaviors of different agents. We provide log features and a helper class for collecting different metrics from the simulation. Furthermore, we implemented a straightforward visual rendering using Pygame to provide an easy-to-view and understand human-readable execution.

\subsection{Reducing the Reality Gap}\label{sec:linkToReality}
The optimal training and evaluation of an RL agent for task offloading hinge on utilizing simulation datasets derived from real-world edge systems. 
Yet, the complexity and diversity inherent to such systems, compounded by the scarcity of standardized reference architectures, benchmarks, and deployment data, present formidable challenges to conducting realistic evaluations of algorithms within actual edge environments~\cite{rausch20hotedge}.
This section delineates our approach to mitigating these challenges by leveraging tools that integrate with PeersimGym to generate plausible topologies~\cite{rausch20hotedge} and workloads~\cite{tian19TraceGenereation}, thereby simulating environments that more closely mirror reality.

\textbf{Trace-generator tool.}
We employ the trace-generation tool to synthesize workloads based on real-world cluster traces from Alibaba Cloud~\cite{tian19TraceGenereation}, producing datasets that mirror actual computational demands. This tool generates workloads comprising multiple jobs, each depicted as a Directed Acyclic Graph (DAG) of tasks, where each task may include several instances requiring specific memory and CPU resources.
While the ubiquity of DAGs in real-world applications is undeniable, the current iteration of our simulator does not support them. Consequently, we interpret the jobs within the trace-generator dataset as necessitating the peak CPU and memory resources identified across all tasks within a job. We calculate task instructions by factoring in the CPU frequency, the requested cores, and task duration. The maximum memory usage across all tasks within a job is considered equivalent to the data size in our simulation.

To incorporate the trace-generated data into PeersimGym, we develop a Python script (\textit{Utils/DatasetGen.py}) processing the dataset and outputs a \textit{JSON} file. This file is then utilized by the AlibabaTraceClient, a client implementation within our simulator, which samples tasks based on the synthesized dataset.
 
\textbf{Topology Generator.}
The \textit{Ether} tool~\cite{rausch20hotedge} enables the generation of realistic infrastructure configurations, drawing from various Edge Computing scenarios. Our focus is on the Urban Sensing Scenario, inspired by the Array of Things project, which emphasizes data collection in smart cities.
This scenario features clusters equipped with sensor nodes, each powered by Single Board Computers (SBCs) and connected to base stations comprising servers and GPU-equipped machines. These nodes process tasks locally or offload them to more capable nodes, simulating a realistic Edge Computing environment.

We augmented the Ether project to facilitate the generation of a topology and the assignment of coordinates to nodes. This setup ensures communication compatibility among nodes within the generated network, allowing for seamless integration with PeersimGym through a helper script that imports the topology data, thereby enhancing the realism of the simulator and applicability to real-world Edge Computing scenarios.

\section{Experimental Results and Analysis}
\label{sec:Exp}
In this section, we evaluate PeersimGym by addressing two pivotal research questions (RQs) that underscore the adaptability and scalability of the MARL solution under varying network conditions:
\begin{description}
\item[RQ1] How does the MARL solution adapt its behavior to a fixed network topology and varying task arrival rates?
\item[RQ2] How does the MARL solution adapt its behavior to a fixed task arrival rate with an increasing number of nodes and agents?
\end{description}

\subsection{Experimental Setup}
The experiments are grounded in four distinct network topologies generated using Ether, each featuring an incremental number of Array of Things (AoT) clusters. These scenarios leverage the realistic topology and workload generation methodologies detailed in Sec.~\ref{sec:linkToReality}. Each AoT cluster, predominantly composed of Single Board Computers (SBCs), includes pairs of SBCs and a base station equipped with an Intel NUC and two GPU units, alongside a remote, more potent server. The simulation parameters for all scenarios can be found in the Agent Repository. 
\begin{figure}[t]
\centering
    \begin{subfigure}{0.32\linewidth}
        \includegraphics[width=\textwidth,height=2.8cm]{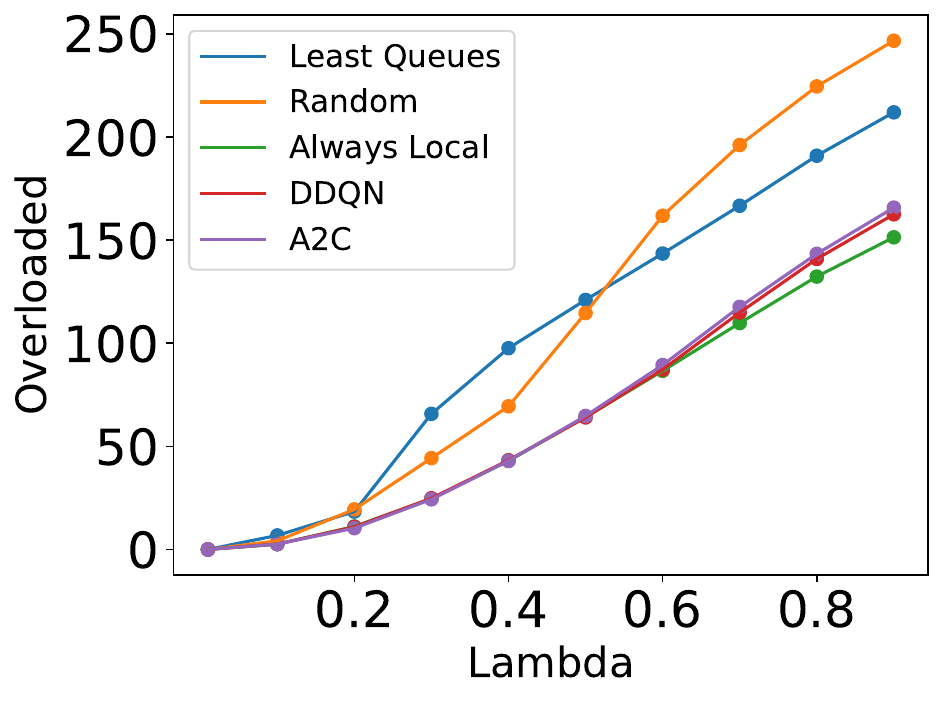}
        \label{fig:overload_results_le}
    \end{subfigure}
    \hfill
    \begin{subfigure}{0.32\linewidth}
        \includegraphics[width=\textwidth,height=2.8cm]{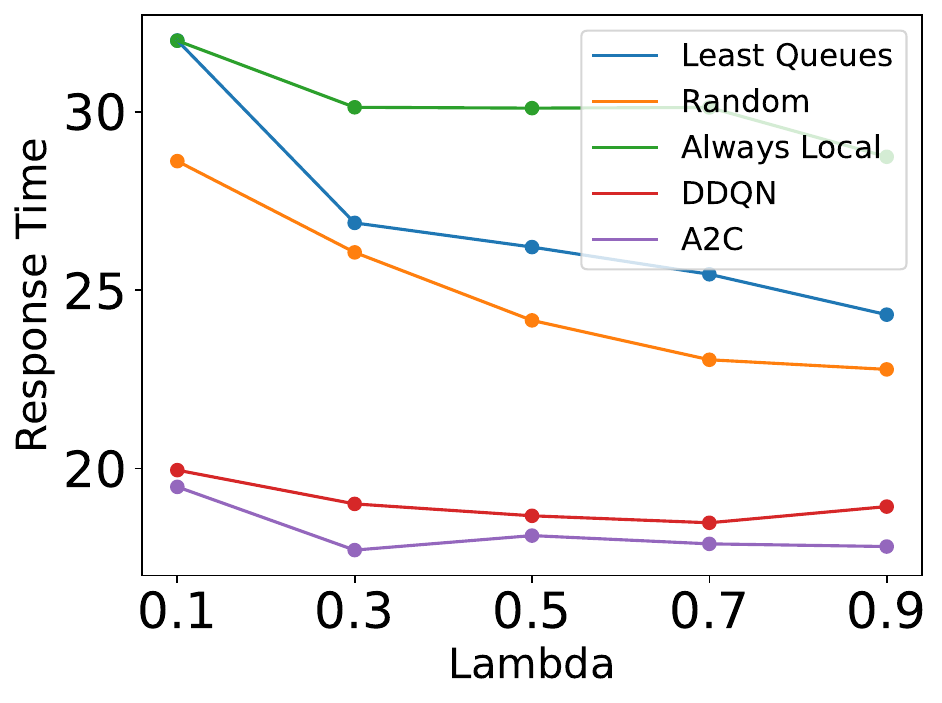}
        \label{fig:response_time_results_le}
    \end{subfigure}
    \hfill
    \begin{subfigure} {0.32\linewidth}  
            \includegraphics[width=\textwidth,height=2.8cm]{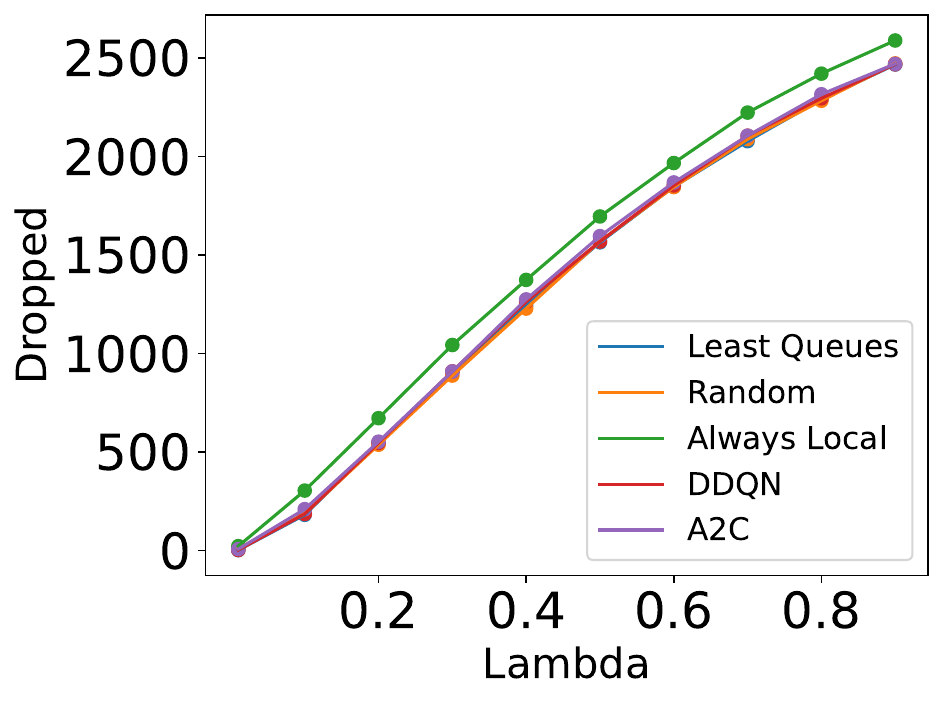}
            \label{fig:dropped_tasks_perc_le}
    \end{subfigure}
 \begin{subfigure}{0.32\linewidth}
        \includegraphics[width=\textwidth, height=2.8cm]{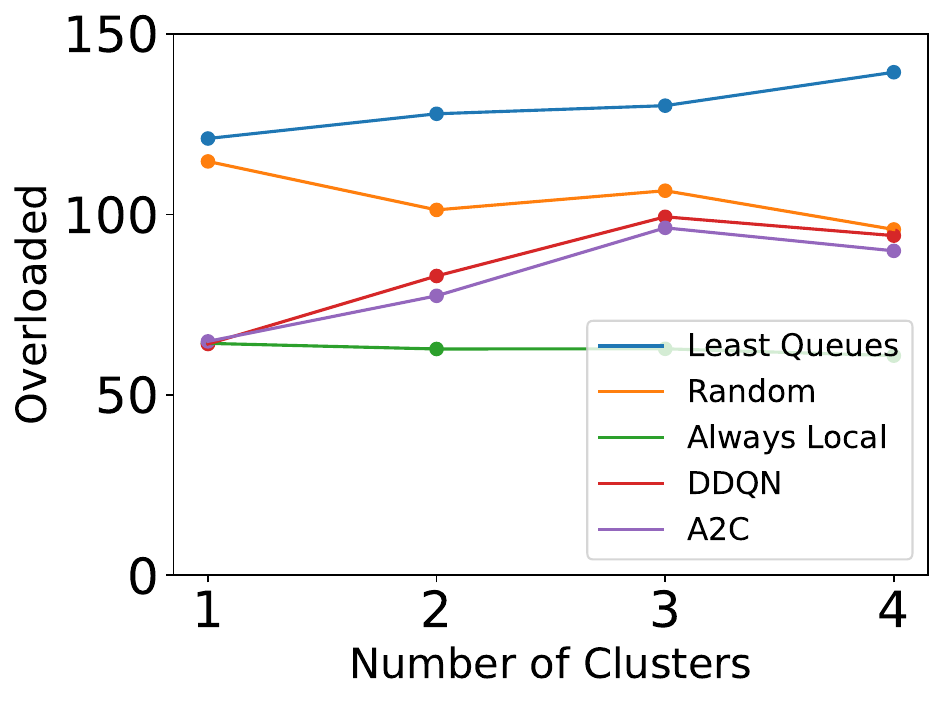}
        \caption{Overloaded nodes}
        \label{fig:overload_results_cluster}
    \end{subfigure}
    \hfill
    \begin{subfigure}{0.32\linewidth}
        \includegraphics[width=\textwidth, height=2.8cm]{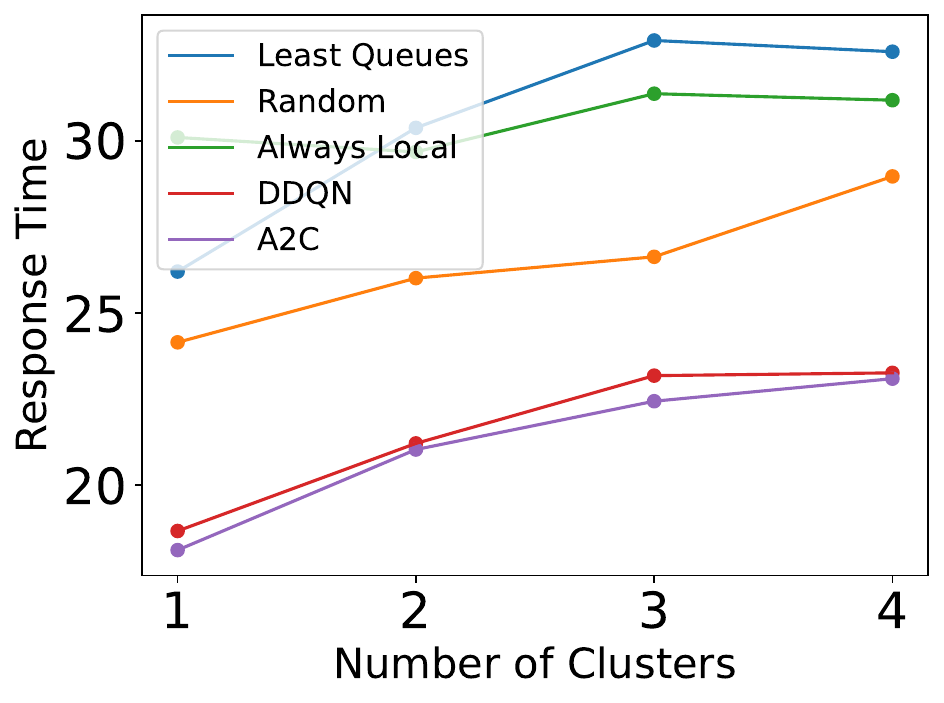}
        \caption{Response time}
        \label{fig:response_time_results_cluster}
    \end{subfigure}
    \hfill
    \begin{subfigure} {0.32\linewidth}  
        \includegraphics[width=\textwidth, height=2.8cm]{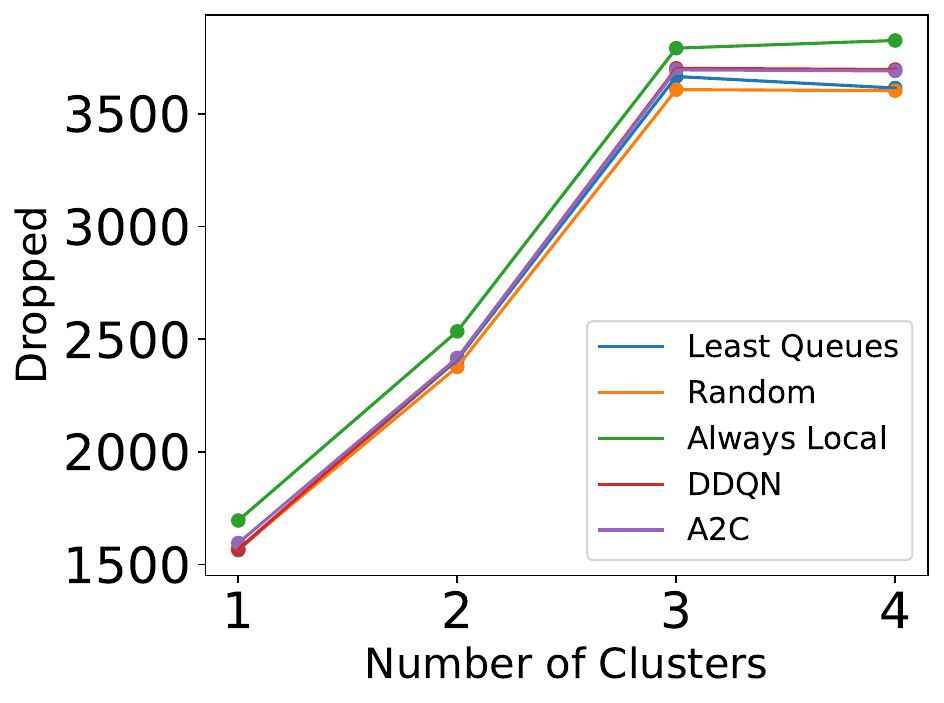}
        \caption{\# of dropped tasks}
        \label{fig:dropped_tasks_perc_cluster}
    \end{subfigure}
    \caption{Evolution of the different metrics with variable $\lambda$ (top) and variable cluster number (bottom), averaging 100 episodes.}
    \label{lambda_variation_plots}
\end{figure}
The topologies vary in cluster numbers, spanning one to four clusters, and correspondingly in node counts, from 12 to 31. The controller protocol is present in varying quantities (8 to 22 nodes across different topologies). Despite uniform CPU frequencies, the computational power varies with the core count, with the remote server having the highest capacity. The communication capabilities and bandwidths are homogeneous across all nodes. Task generation at each SBC node follows a $Poisson(\lambda)$ distribution, over a simulation episode of $1000$ time steps. Each agent makes an offloading decision in each time step, through $300$ training and $100$ inference episodes. See Tab.~\ref{tab:param} for exact values \footnote{We provide the in-depth configurations for the environment in the agent repository}
\begin{table}[t]
    \centering
    \caption{Parameter values in the experimental setup.}
    \begin{minipage}[t]{0.48\linewidth}
        \centering
        \begin{tabular}{r|l}
            \hline
            \hline
               {\color{white}..............} Simulation time, $T$ & 1000 s\\
        \hline
                   Task arrival rate, $\lambda$ & 0.17 \\
        Task input size, $\alpha_i^{\text{in}}$ & 150 Mbytes\\
      Task output size, $\alpha_i^{\text{out}}$ & 150 Mbytes\\
                    Task instructions, $\rho_i$ & 8e7 \\
                                   CPI, $\xi_i$ & 1 \\
                           Deadline, $\delta_i$ & 100 s\\
            \hline
                Bandwidth, $B_{i,j}$ & 2 MHz \\
           Transmission power, $P_i$ & 20 dbm\\
        \hline
        \hline        
        \end{tabular}
    \end{minipage}%
    \hfill
    \begin{minipage}[t]{0.48\linewidth}
        \centering
        \begin{tabular}{r|l}
            \hline
            \hline
                               Nodes per tier & (10, 10, 10) \\
               Processor frequencies $\phi_i$ & $(4, 2, 8)e^7$ MHz\\
                Numbers of cores $N_{\phi}^i$ & (1, 1, 2) \\
        Buffer capacities, $Q_i^{\text{max}}$ & (20, 10, 100)\\
        \hline
                               Task utility, $r_u$ & 2 \\
            Weight waiting, $\chi_D^{\text{wait}}$ & 20 \\
          Weight excecution, $\chi_D^{\text{exc}}$ & 20 \\
               Weight comm, $\chi_D^{\text{comm}}$ & 20 \\
                         Weight overload, $\chi_O$ & 150 \\
            \hline
            \hline        
        \end{tabular}
    \end{minipage}
    \label{tab:param} 
\end{table}

To answer the RQs, we test the performance of the DDQN and A2C agents and a set of baseline approaches:
    1.) \textit{Local Processing} -- never offload tasks. 
    2.) \textit{Random Offloading} -- select the target node randomly.
    3.) \textit{Least Queue} -- select the neighbor with the shortest task queue at the time as a target.

\textbf{Metrics.}
To show that our simulation can be used effectively to train RL agents in a MARL fashion, we select three commonly used metrics in the field to test the agent: 1.) the number of times a node is overloaded, 2.) the average task completion time for the tasks, and 3.) the number of dropped tasks. 

\subsection{Results Analysis}
\textbf{RQ1:}
We examine the behaviors of different agents when faced with different workloads by varying the parameter $\lambda$ that governs the task arrival rate of each SBC node. In Fig. \ref{lambda_variation_plots} (top), we observe that, with increasing $\lambda$, the nodes run out of available resources quickly. The almost linear increase in dropped tasks and overloaded nodes could indicate that the nodes are already working at capacity even for $\lambda\approx 0.1$. The reduction in response time observed in the \textit{Least Queue} and \textit{Random} approaches could be explained by more tasks being processed locally, due to the influx of tasks arriving surpassing the number of tasks offloaded keeping the node always full. In terms of task dropping and node overloading, the \textit{DDQN} constantly achieves as reasonable rates as the best baselines. At the same time, it gives significantly better response time, with a large margin of difference, and apparently not affected by the task arrival rate. Behavior of \textit{A2C} follows closely that of \textit{DDQN}.

\textbf{RQ2:} Effects of increasing the number of nodes are shown in Fig.~\ref{lambda_variation_plots} (bottom). We test four network topologies of the AoT scenario and observe that the number of dropped tasks increases almost linearly with the increase in the number of nodes -- note that the sharp break at the end is due to the 3-cluster topology having nearly the same number of nodes as the 4-cluster topology. This can indicate that the network is exhausting its resources for all of the topologies. The number of dropped tasks increases because the number of nodes, and therefore the total amount of tasks generated, increases. The response times also observe a slight increase due to the more powerful server that is shared across all the clusters filling up, and therefore, the tasks spend more time waiting in its queue, reducing the concurrency that was available when fewer nodes could access the server. The number of overloaded nodes also increases for the \textit{Least Queue}, \textit{A2C}, and \textit{DDQN}, since the shared server is overloading more often. The number of overloads for the \textit{Local Processing} remains stable and is lower than the other approaches because it never incurs the cost of overloading the base station or the server. Still, in turn, the number of dropped tasks is also higher. RL solutions again show clear superiority in terms of response times compared to baselines, while not sacrificing other metrics.

\section{Conclusion and Future Work}
We introduced PeersimGym, a highly customizable environment for the development and evaluation of MARL-based solutions to the task offloading problem in Edge Computing systems. 
PeersimGym integrates a MARL environment, compatible with the PettingZoo framework, facilitating agent interaction within a Python-based setting, and a simulation platform constructed atop Peersim, a Java tool for simulating Peer-to-Peer (P2P) networks. 
Our framework encompasses a suite of protocols for controllers, workers, and networking layers, offering the flexibility to modify system behaviors through configuration changes. The demonstrated efficacy of PeersimGym across various network settings highlights its potential as a training ground for MARL strategies in edge environments.
To foster further research and collaboration within the community, we have made the source code for both the \href{https://github.com/FredericoMetelo/peersim-environment}{simulation environment} and the \href{https://github.com/FredericoMetelo/TaskOffloadingAgentLibrary}{agent development toolkit} publicly available. Our documentation, including comprehensive wikis, provides detailed guidance on utilizing PeersimGym, encouraging researchers to explore and benchmark their own RL algorithms using our tool. This initiative aims to accelerate advancements in the domain and contribute to the broader efforts of the research community in optimizing Edge Computing systems.

\textbf{Future Enhancements.}
Future enhancements for PeersimGym will focus on dynamic Edge Computing features: (1) Node and System Mobility, adding mobile node and service models with replicas for uninterrupted service; (2) Task Diversity, introducing a wider range of tasks tailored to specific hardware capabilities; (3) Data Locality Sensitivity, implementing prioritization based on data proximity to streamline offloading; and (4) Federated Reinforcement Learning, adopting collaborative training methods across nodes to optimize learning and parameter sharing.

\textbf{Broader Impact.}
The versatility of PeersimGym spans beyond Edge Computing into fields like smart grids and satellite communications, where it can optimize energy management and enhance connectivity, respectively. Its adaptability to various P2P scenarios demonstrates the potential for broad technological impacts, though some domain-specific modifications may be required.

\textbf{Acknowledgements} This work was partially supported by NOVA LINCS (UI DB/04516/2020) with the financial support of FCT I.P. and Project "Artificial Intelligence Fights Space Debris" No C626449889-0046305 co-funded by Recovery and Resilience Plan and NextGeneration EU Funds, www.recuperarportugal. gov.pt.
And, by the European Union (TARDIS, 101093006). Views and opinions expressed are however those of the author(s) only and do not necessarily reflect those of the European Union. Neither the European Union nor the granting authority can be held responsible for them.

\bibliographystyle{unsrt}
\bibliography{references} 

\begin{thebibliography}{10}

\bibitem{min2019learning}
Minghui Min, Liang Xiao, Ye~Chen, Peng Cheng, Di~Wu, and Weihua Zhuang.
\newblock Learning-based computation offloading for iot devices with energy
  harvesting.
\newblock {\em IEEE Transactions on Vehicular Technology}, 68(2):1930--1941,
  2019.

\bibitem{Muniswamaiah_Manoj_2021}
Manoj Muniswamaiah, Tilak Agerwala, and Charles~C. Tappert.
\newblock A survey on cloudlets, mobile edge, and fog computing.
\newblock In {\em 8th IEEE CSCloud/7th IEEE EdgeCom}, 2021.

\bibitem{Varghese_Buyya_2018}
Blesson Varghese and Rajkumar Buyya.
\newblock Next generation cloud computing: New trends and research directions.
\newblock {\em Future Generation Computer Systems}, 79:849--861, 2018.

\bibitem{yu2020deep}
Shuai Yu, Xu~Chen, Zhi Zhou, Xiaowen Gong, and Di~Wu.
\newblock When deep reinforcement learning meets federated learning:
  Intelligent multitimescale resource management for multiaccess edge computing
  in 5g ultradense network.
\newblock {\em IEEE Internet of Things Journal}, 8(4):2238--2251, 2020.

\bibitem{zhu2019blot}
Zhaowei Zhu, Ting Liu, Yang Yang, and Xiliang Luo.
\newblock Blot: Bandit learning-based offloading of tasks in fog-enabled
  networks.
\newblock {\em IEEE Trans. Parallel Distrib. Syst.}, 2019.

\bibitem{lin2023deep}
Lixia Lin, Wen Zhou, Zhicheng Yang, and Jianlong Liu.
\newblock Deep reinforcement learning-based task scheduling and resource
  allocation for noma-mec in industrial internet of things.
\newblock {\em Peer-to-Peer Networking and Applications}, 16(1):170--188, 2023.

\bibitem{zhang2023cooperative}
Fan Zhang, Guangjie Han, Li~Liu, Yu~Zhang, Yan Peng, and Chao Li.
\newblock Cooperative partial task offloading and resource allocation for iiot
  based on decentralized multi-agent deep reinforcement learning.
\newblock {\em IEEE Internet of Things Journal}, 2023.

\bibitem{peersim}
Alberto Montresor and M\'{a}rk Jelasity.
\newblock {PeerSim}: A scalable {P2P} simulator.
\newblock In {\em Proc. of the 9th Int. Conference on Peer-to-Peer}, pages
  99--100, Seattle, WA, September 2009.

\bibitem{PettingZoo_2020}
Justin~K. Terry, Benjamin Black, Ananth Hari, Luis~S. Santos, Clemens
  Dieffendahl, Niall~L. Williams, Yashas Lokesh, Caroline Horsch, and Praveen
  Ravi.
\newblock Pettingzoo: Gym for multi-agent reinforcement learning.
\newblock {\em CoRR}, abs/2009.14471, 2020.

\bibitem{qiu2019online}
Xiaoyu Qiu, Luobin Liu, Wuhui Chen, Zicong Hong, and Zibin Zheng.
\newblock Online deep reinforcement learning for computation offloading in
  blockchain-empowered mobile edge computing.
\newblock {\em IEEE Transactions on Vehicular Technology}, 68(8):8050--8062,
  2019.

\bibitem{baek2019managing}
Jung-yeon Baek, Georges Kaddoum, Sahil Garg, Kuljeet Kaur, and Vivianne Gravel.
\newblock Managing fog networks using reinforcement learning based load
  balancing algorithm.
\newblock In {\em 2019 IEEE WCNC}, pages 1--7. IEEE, 2019.

\bibitem{Baek22Fog}
Jungyeon Baek and Georges Kaddoum.
\newblock Floadnet: Load balancing in fog networks with cooperative multiagent
  using actor-critic method.
\newblock {\em IEEE Trans. Netw. Serv. Manage.}, 2023.

\bibitem{van2018deep}
Duc Van~Le and Chen-Khong Tham.
\newblock A deep reinforcement learning based offloading scheme in ad-hoc
  mobile clouds.
\newblock In {\em IEEE INFOCOM WKSHPS}, pages 760--765, 2018.

\bibitem{dai2022task}
Fei Dai, Guozhi Liu, Qi~Mo, WeiHeng Xu, and Bi~Huang.
\newblock Task offloading for vehicular edge computing with edge-cloud
  cooperation.
\newblock {\em World Wide Web}, 25(5):1999--2017, 2022.

\bibitem{peng2022deep}
Xin Peng and et~al.
\newblock Deep reinforcement learning for shared offloading strategy in vehicle
  edge computing.
\newblock {\em IEEE Systems Journal}, 2022.

\bibitem{liu2019deep}
Yi~Liu, Huimin Yu, Shengli Xie, and Yan Zhang.
\newblock Deep reinforcement learning for offloading and resource allocation in
  vehicle edge computing and networks.
\newblock {\em IEEE Transactions on Vehicular Technology}, 68(11):11158--11168,
  2019.

\bibitem{geng2023deep}
Liwei Geng, Hongbo Zhao, Jiayue Wang, Aryan Kaushik, Shuai Yuan, and Wenquan
  Feng.
\newblock Deep reinforcement learning based distributed computation offloading
  in vehicular edge computing networks.
\newblock {\em IEEE Internet of Things Journal}, 2023.

\bibitem{huang2021deadline}
Hui Huang, Qiang Ye, and Yitong Zhou.
\newblock Deadline-aware task offloading with partially-observable deep
  reinforcement learning for multi-access edge computing.
\newblock {\em IEEE Transactions on Network Science and Engineering},
  9(6):3870--3885, 2021.

\bibitem{jain2023qos}
Vibha Jain and Bijendra Kumar.
\newblock Qos-aware task offloading in fog environment using multi-agent deep
  reinforcement learning.
\newblock {\em J. Netw. Syst. Manag.}, 31(1):7, 2023.

\bibitem{tong2023multi}
Zhao Tong, Jiake Wang, Jing Mei, Kenli Li, Wenbin Li, and Keqin Li.
\newblock Multi-type task offloading for wireless internet of things by
  federated deep reinforcement learning.
\newblock {\em Future Generation Computer Systems}, 145:536--549, 2023.

\bibitem{Somnez_2018_EdgeCloudSim}
Cagatay Sonmez, Atay Ozgovde, and Cem Ersoy.
\newblock Edgecloudsim: An environment for performance evaluation of edge
  computing systems.
\newblock {\em Transactions on Emerging Telecommunications Technologies},
  29(11):e3493, 2018.
\newblock e3493 ett.3493.

\bibitem{Redowan_Mahmud_2021_ifogsim2}
Md.~Redowan Mahmud, Samodha Pallewatta, Mohammad Goudarzi, and Rajkumar Buyya.
\newblock Ifogsim2: An extended ifogsim simulator for mobility, clustering, and
  microservice management in edge and fog computing environments.
\newblock {\em CoRR}, abs/2109.05636, 2021.

\bibitem{Boni_Hassan_Drira_2021_ns3gym}
Piotr Gaw{\l}owicz and Anatolij Zubow.
\newblock {ns-3 meets OpenAI Gym: The Playground for Machine Learning in
  Networking Research}.
\newblock In {\em {ACM International Conference on Modeling, Analysis and
  Simulation of Wireless and Mobile Systems (MSWiM)}}, 2019.

\bibitem{Santos_Wauters_2020_FogGym}
Jos{\'e} Santos, Tim Wauters, Bruno Volckaert, and Filip De~Turck.
\newblock Reinforcement learning for service function chain allocation in fog
  computing.
\newblock {\em Book Chapter in revision, Submitted to Communications Network
  and Service Management In the Era of Artificial Intelligence and Machine
  Learning, IEEE Press}, 2020.

\bibitem{towers_gymnasium_2023}
Mark Towers and et~al.
\newblock Gymnasium, March 2023.

\bibitem{mnih2015human}
Volodymyr Mnih, Koray Kavukcuoglu, David Silver, Andrei~A Rusu, Joel Veness,
  Marc~G Bellemare, Alex Graves, Martin Riedmiller, Andreas~K Fidjeland, Georg
  Ostrovski, et~al.
\newblock Human-level control through deep reinforcement learning.
\newblock {\em Nature}, 518(7540):529--533, 2015.

\bibitem{Nowé2012}
Ann Now{\'e}, Peter Vrancx, and Yann-Micha{\"e}l De~Hauwere.
\newblock {\em Game Theory and Multi-agent Reinforcement Learning}.
\newblock Springer Berlin Heidelberg, Berlin, Heidelberg, 2012.

\bibitem{ng1999rewardshaping}
Andrew~Y Ng, Daishi Harada, and Stuart Russell.
\newblock Policy invariance under reward transformations: Theory and
  application to reward shaping.
\newblock In {\em Icml}, volume~99, pages 278--287, 1999.

\bibitem{book:Anttalainen_2003}
Tarmo Anttalainen.
\newblock {\em Introduction to telecommunications network engineering}.
\newblock Artech House telecommunications library. Artech House, Boston, 2nd
  edition, 2003.

\bibitem{rausch20hotedge}
Thomas Rausch, Clemens Lachner, Pantelis~A Frangoudis, Philipp Raith, and
  Schahram Dustdar.
\newblock Synthesizing plausible infrastructure configurations for evaluating
  edge computing systems.
\newblock In {\em 3rd USENIX Workshop HotEdge 20}, 2020.

\bibitem{tian19TraceGenereation}
Huangshi Tian, Yunchuan Zheng, and Wei Wang.
\newblock Characterizing and synthesizing task dependencies of data-parallel
  jobs in alibaba cloud.
\newblock In {\em Proc. ACM Symp. Cloud Comput.}, 2019.

\end{thebibliography}

\end{document}